\documentclass[letterpaper, 10 pt, conference]{ieeeconf}

\IEEEoverridecommandlockouts                            
\usepackage{graphicx}
\usepackage{balance}
\usepackage{comment}
\usepackage{cite}
\usepackage{amsmath, amssymb}
\usepackage[tight,footnotesize]{subfigure}
\usepackage[active]{srcltx}
\usepackage{mathtools}
\graphicspath{{./figs/}}
\renewcommand{\baselinestretch}{1}
\usepackage{algorithm}
\usepackage{algpseudocode}
\usepackage{multicol}
\usepackage{dsfont}
\usepackage{amsfonts}
\usepackage{cases}
\usepackage{bm} 

\usepackage{fancyhdr}
\fancypagestyle{preprintfirstpage}{%
  \fancyhf{}%
  \fancyhead[L]{\footnotesize\bfseries 2025 IEEE 64th Conference on Decision and Control (CDC)\\December 10-12, 2025. Rio de Janeiro, Brazil\\(extended version)}%
}

\newcommand{\ct}{\cos\left(\theta(i)\right)}
\newcommand{\st}{\sin\left(\theta(i)\right)}
\newcommand{\U}{\text{x}^i_s \ct - \text{y}^i_s \st}
\newcommand{\V}{\text{x}^i_s \st + \text{y}^i_s \ct}

\begin{document}

\title{\LARGE \bf
Adaptive Monitoring of Stochastic Fire Front Processes via Information-seeking Predictive Control
\vspace{-0mm}}

\author{Savvas~Papaioannou,~Panayiotis~Kolios,~Christos~G.~Panayiotou and~Marios~M.~Polycarpou
\thanks{The authors are with the KIOS Research and Innovation Centre of Excellence (KIOS CoE) and the Department of Electrical and Computer Engineering, University of Cyprus, Nicosia, 1678, Cyprus. E-mail:{\tt\small \{papaioannou.savvas, pkolios, christosp, mpolycar\}@ucy.ac.cy}%
\newline
This work is supported by the European Union's Horizon Europe program under grant agreement No 101187121 (EUSOME) and the Civil Protection Knowledge for Action in Prevention \& Preparedness under grant agreement No. 101193719 (COLLARIS2). It is also supported from the Republic of Cyprus through the Deputy Ministry of Research, Innovation and Digital Policy.
}}

\maketitle

\thispagestyle{preprintfirstpage}

\begin{abstract}
We consider the problem of adaptively monitoring a wildfire front using a mobile agent (e.g., a drone), whose trajectory determines where sensor data is collected and thus influences the accuracy of fire propagation estimation.
This is a challenging problem, as the stochastic nature of wildfire evolution requires the seamless integration of sensing, estimation, and control, often treated separately in existing methods. State-of-the-art methods either impose linear-Gaussian assumptions to establish optimality or rely on approximations and heuristics, often without providing explicit performance guarantees. To address these limitations, we formulate the fire front monitoring task as a stochastic optimal control problem that integrates sensing, estimation, and control. We derive an optimal recursive Bayesian estimator for a class of stochastic nonlinear elliptical-growth fire front models. Subsequently, we transform the resulting nonlinear stochastic control problem into a finite-horizon Markov decision process and design an information-seeking predictive control law obtained via a lower confidence bound-based adaptive search algorithm with asymptotic convergence to the optimal policy.
\end{abstract}

\section{Introduction} \label{sec:Introduction}
 
 The 2025 Southern California wildfires were among the most devastating in the state's history: the Palisades Fire in Los Angeles and the Eaton Fire in Altadena burned nearly 40,000 acres, destroyed over 16,000 structures, and displaced hundreds of thousands of people~\cite{Kajita2025notes}. Accurate estimation of wildfire propagation is therefore critical for effective disaster response \cite{papaioannou2021towards,papaioannou2020cooperative,papaioannou2020coordinated} and informed decision-making \cite{papaioannou20213d,papaioannou2024synergising}. Motivated by this need, this work investigates adaptive monitoring of a stochastic wildfire front using a mobile agent.  

The task requires planning the agent's trajectory over a rolling finite horizon to minimize uncertainty in estimating the fire's evolution from sensor data. At each step, the agent re-plans based on the current environment, yielding a complex problem that demands an integrated approach to sensing \cite{papaioannou2019jointly,papaioannou2023cooperative,papaioannou2022integrated}, estimation\cite{Papaioannou2019decentralized,papaioannou2023joint,papaioannou2022distributed} , and control \cite{papaioannou2025rolling,papaioannou2025jointly,11186919}. Existing methods often address only parts of this problem, either by decoupling sensing, estimation , and control, or by simplifying assumptions~\cite{Xuan2020,Sujit2007cooperative}. 
 The problem considered in this work relates to \textit{active sensing}~\cite{Lauri2014stochastic}, \textit{information gathering}~\cite{Atanasov2014information}, and \textit{sensor management}~\cite{Hero2011sensor}. Sensor management approaches typically address stateless sensors without dynamics, focusing on placement or node selection, and are therefore limited in scenarios where sensor states evolve (e.g., drones equipped with onboard sensors). In such dynamic settings, adaptive control strategies are required. For stochastic dynamic processes, many methods adopt myopic control~\cite{Dames2012decentralized}, whereas non-myopic schemes are typically greedy~\cite{Singh2009efficient} or heuristic with sub-optimality guarantees~\cite{Hollinger2014sampling}. Approaches that provide optimality guarantees often assume linear dynamics with Gaussian noise and apply the certainty equivalence principle (CEP), thereby reducing the problem to deterministic optimal control~\cite{Kantaros2021sampling}. These assumptions, however, do not hold for the problem addressed in this work.

In summary, this work integrates sensing, estimation, and control within a unified stochastic optimal control (SOC) framework for adaptive wildfire-front monitoring using a mobile agent. We develop a recursive Bayesian estimator for elliptical fire-front dynamics under limited sensing and uncertainty, and reformulate the nonlinear SOC problem as a finite-horizon Markov decision process (MDP). The MDP is solved via a lower-confidence-bound (LCB) guided adaptive search that asymptotically converges to the optimal policy.

The rest of the paper is organized as follows: Section \ref{sec:problem} formulates the problem, Section \ref{sec:approach} presents the proposed methodology, Section \ref{sec:Evaluation} provides an evaluation of the approach, and finally, Section \ref{sec:conclusion} concludes the paper.

\section{Problem Formulation} \label{sec:problem}

\subsection{Problem Objective} \label{ssec:formulation}

Let a mobile agent be at state \( y_t \) at time step \( t \), and let the agent's belief distribution over the fire front state \( X_t \) be denoted by $\mathcal{B}_t(X_t|Z_{1:t})$ (abbreviated as $\mathcal{B}_t$) which was computed using measurements $Z_{1:t}=[Z_1,\ldots,Z_t]$ up to time step $t$. The objective is to compute the optimal sequence of control inputs \( \{u_{t|t}, \ldots, u_{t+T-1|t}\} \) over a finite rolling planning horizon of \( T \) time steps that minimizes:
\begin{equation}
	\mathbb{E}_{Z_t^{1:T}} \left\{ \sum_{\tau=1}^{T} \mathcal{C}_{\tau} \left( \mathcal{B}^{-}_{t+\tau|t}, Z_{t+\tau|t}(u_{t+\tau-1|t}) \right) \right\}.
\end{equation}
\noindent Here, the subscript \( t+\tau|t \) denotes predicted quantities at time step \( t+\tau \) for \( \tau \in \{1, \dots, T\} \) within the planning horizon, based on information available at time step \( t \). The cost function \( \mathcal{C}_\tau \) is a bounded, real-valued function that takes as input: a) the predictive density $\mathcal{B}^{-}_{t+\tau|t}$ i.e., $\mathcal{B}^{-}_{t+\tau|t}(X_{t+\tau|t}|Z_{1:t+\tau-1|t})$, of the fire front state at time \( t+\tau|t \), and b) the future i.e., predicted, measurement \( Z_{t+\tau|t}(u_{t+\tau-1|t}) \), which will be received given that the agent executes the control input \( u_{t+\tau-1|t} \) and moves to the predicted state $y_{t+\tau|t}$. It then returns a value representing the uncertainty of the fire front state captured in the resulting (pseudo) posterior belief \( \mathcal{B}_{t+\tau|t} \). 

An information-rich measurement set is one that reduces the dispersion in the posterior, which is the agent's objective over the planning horizon. The expectation is taken with respect to the future measurement set \( Z_t^{1:T} = \{Z_{t+1|t}, \ldots, Z_{t+T|t}\} \). Subsequently, the agent executes the first control input in the sequence, i.e., \( u_{t|t} \), transitions to its new state \( y_{t+1} \), receives the real measurement \( Z_{t+1} \), computes the posterior belief \( \mathcal{B}_{t+1}(X_{t+1} | Z_{1:t+1}) \), and repeats the process described above for time step $t+1$.

\subsection{Fire Front Propagation Model} \label{ssec:fire_model}

In this work, as we discuss next, we employ a stochastic adaptation of the deterministic elliptical fire propagation model proposed in \cite{Richards1990elliptical}. This model, which is currently utilized in various fire-area simulators \cite{Finney1998farsite}, describes the spatiotemporal evolution of a fire front using a nonlinear system of first-order differential equations. Specifically, the fire front is represented as an ellipse defined by a series of $N$ vertices that collectively delineate the propagating fire's edge at a specific moment. The spatiotemporal discrete-time dynamics of vertex $i \in \{1,\ldots,N\}$ at time step $t$ are given by:
\begin{equation}\label{eq:fire_model_vertex}
	x^i_{t} = x^i_{t-1} + \Delta t \, \dot{x}^i_{t-1},
\end{equation}
\noindent where $x^i_{t} = [\text{x}^i_t,\text{y}^i_t]^\top \in \mathbb{R}^2$ is the state of vertex $i$ composed of 2D Cartesian coordinates, $\Delta t$ is the sampling interval, and the fire growth velocity at vertex $i$ is given by $\dot{x}^i_{t-1}=$
\begin{equation}
\begin{bmatrix}
\displaystyle
\frac{\alpha_1^2(i) \ct SC(i) - \alpha_2^2(i) \st CS(i)}{\sqrt{\alpha_2^2(i) CS(i)^2 + \alpha_1^2(i) SC(i)^2}} + C_1(i) \!\\
\displaystyle
\frac{-\alpha_1^2(i) \st SC(i) - \alpha_2^2(i) \ct CS(i)}{\sqrt{\alpha_2^2(i) CS(i)^2 + \alpha_1^2(i) SC(i)^2}} + C_2(i)\!
\end{bmatrix},\notag
\end{equation}

\noindent where:

\begin{equation}
\begin{aligned}
SC(i) &= \V, \\
CS(i) &= \U, \\
C_1(i) &= \alpha_3(i)\,\st, \\
C_2(i) &= \alpha_3(i)\,\ct, \\
\end{aligned}
\end{equation}

\noindent and $[\text{x}^i_s,\text{y}^i_s]^\top$  are the components of the tangent vector at vertex $i$, providing the local orientation of the fire front at that point. 

Environmental conditions, such as fuel type and weather, local to each vertex, affect the forward fire propagation rate and direction. These factors include wind direction and speed, denoted by $\theta$ and $w_s$, respectively, as well as the fire spread rate due to fuel type, denoted by $r_f$. These parameters are stochastic and can vary throughout the environment. Therefore, each vertex may be affected differently depending on the fire front's extent and the environmental variability. 

Specifically, we denote $\theta(i) \in [0, 2\pi]$ as the wind direction affecting vertex $i$, the wind speed at the location of vertex $i$ as $w_s(i) \in \mathbb{R}^+$, and the fire spread rate as $r_f(i) \in \mathbb{R}^+$. Here, $\theta(i), w_s(i), r_f(i)$, $\forall i$, are random realizations of the wind direction, wind speed, and fire spread rate at the location of vertex $i$.

The parameters $\alpha_1(i)$, $\alpha_2(i)$, and $\alpha_3(i)$ denote the shape parameters governing the elliptical fire growth from vertex $i$, representing respectively the lengths of the semi-minor axis, the semi-major axis, and the distance from the ignition point to the center of the ellipse, defined respectively as:
\begin{equation}
\begin{aligned}
\alpha_1(i) &= \frac{r_f(i) + \dfrac{r_f(i)}{HB(i)}}{2\,LB(i)}, \\
\alpha_2(i) &= \frac{r_f(i) + \dfrac{r_f(i)}{HB(i)}}{2}, \\
\alpha_3(i) &= \alpha_2(i) - \frac{r_f(i)}{HB(i)} .
\end{aligned}
\end{equation}

\noindent where $HB(i)$ is the head-to-back ratio accounting for the difference between the fire's forward (head) and backward (back) spread from the ignition point, while $LB(i)$ is the length-to-breadth ratio which determines the overall elongation of the fire's elliptical shape. These are defined as: 
 \begin{equation}
\begin{aligned}
HB(i) &= \frac{LB(i) + \bigl(LB(i)^2 - 1\bigr)^{0.5}}{LB(i) - \bigl(LB(i)^2 - 1\bigr)^{0.5}}, \\
LB(i) &= 0.936\,\exp\!\bigl(0.2566\, w_s(i)\bigr)\\
      &+ 0.461\,\exp\!\bigl(-0.1548\, w_s(i)\bigr)
       - 0.397 .
\end{aligned}
\end{equation}
 \noindent For a more in-depth description these parameters we refer the reader to \cite{Finney1998farsite}. Subsequently, the propagation of the fire front process $X_t = [x^1_t,\ldots,x^N_t]^\top \in \mathcal{X}$ is more compactly expressed as: 
 \begin{equation}
 	X_t = \xi(X_{t-1},E_{t-1}) ,
 \end{equation}
 \noindent  where  \( E_t \sim P_E \), with \( E_t \in [0, 2\pi]^N \times [0, \infty)^N \times [0, \infty)^N \), denotes a random realization of \( \{\theta(i), w_s(i), r_f(i)\}_{i=1}^N \) drawn from the PDF \( P_E \), which captures the stochasticity of the fire front propagation acting as a stationary process noise.

\subsection{Agent Dynamics and Sensing Model} \label{ssec:agent_model}

An autonomous mobile agent represented by a point-mass object, evolves inside a bounded planar environment $\mathcal{E}\subset \mathbb{R}^2$ according to discrete-time dynamics of the form \cite{9836051,papaioannou2023unscented}:
\begin{equation}
	y_t = f_a(y_{t-1}, u_{t-1})
\end{equation}
\noindent where \( y_t \in \mathcal{Y}\) is the state of agent at time \( t \), and \( u_{t} \in \mathcal{U} \) is the control input. In addition, the agent has a finite sensing range for observing its surroundings (i.e., through a camera), which is given by a circular region with radius $R_a$ i.e., $O_t = \{x \in \mathbb{R}^2 \mid \|x - y^p_t\| \leq R_a\}$, where $y^p_t$ is the agent's position at time $t$. 

The agent uses its camera to observe the state of the fire front, i.e., by taking snapshots and determining the location of the fire front from the image snapshots using image processing (i.e., object detection). Due to sensing and image processing imperfections, this process carries a certain degree of inaccuracy, resulting in noisy observations. Specifically, for fire front vertex \( i \) with true state \(x^i_t\), the agent observes the measurement \( z^i_t \) inside its sensing range according to:
\begin{equation}
	 z^i_t = h(x^i_t) + w^i_t ,
\end{equation}
\noindent where \( h(\cdot) \) is a function that relates the true states to the received measurements, and \( w^i_t \sim \mathcal{N}(0,\sigma^2_z I_{2\times2}) \) represents measurement noise. The noise is independent and identically distributed (i.i.d.) according to a zero-mean Gaussian distribution with variance \( \sigma^2_z \), where \( I_{2\times2} \) is the \( 2 \times 2 \) identity matrix. Additionally, \( w^i_t \) is independent of the process noise described in the previous section.

The object detection algorithm often produces multiple fragmented pixel blobs for the same fire-front vertex, so that $x^i_t$ is associated with the measurement vector $[z^{i,1}_t,\ldots,z^{i,n^i_t}_t]$. The number of such blobs depends on the true vertex location, and is modeled as a Poisson random variable with rate $\lambda^i_t(x^i_t)$. Thus, a vertex $x^i_t$ may yield $n^i_t \sim \text{Pois}(\lambda^i_t(x^i_t))$ detections within the sensing range. The resulting measurement set follows a Poisson point process~\cite{Streit2010poisson} with intensity
\begin{equation} \label{eq:PPP}
 	\gamma^i_t(z|x^i_t,y_t) = \lambda^i_t(x^i_t)\, p(z|x^i_t), \quad x^i_t \in O_t,
\end{equation}
and $\gamma^i_t(z|x^i_t,y_t)=0$ otherwise, where $p(z|x^i_t)=\mathcal{N}(z;h(x^i_t),\sigma^2_z I_{2\times2})$ is the normalized measurement likelihood restricted to $O_t$.

\section{Adaptive Monitoring via Information-seeking Predictive Control}\label{sec:approach}

\subsection{Fire Front Recursive State Estimation} \label{ssec:Estimation}

Bayesian recursive state estimation for systems with fixed-dimensional state and measurement vectors is formulated through the predictor-corrector recursion:
\begin{align} \label{eq:Bayes_recursion}
& \mathcal{B}^{-}_t(x_t|z_{1:t-1}) = \int p(x_t|x_{t-1})\, \mathcal{B}_{t-1}(x_{t-1}|z_{1:t-1}) \,dx_{t-1}, \notag \\
& \mathcal{B}_t(x_t|z_{1:t}) = \frac{p(z_t|x_t) \, \mathcal{B}^{-}_t(x_t|z_{1:t-1})}{\int p(z_t|x_t) \,\mathcal{B}^{-}_t(x_t|z_{1:t-1}) \, dx_t}, 
\end{align}

\noindent where with slight abuse of notation in Eq. \eqref{eq:Bayes_recursion}  $x_t \in \mathbb{R}^{d_x}$ is the state of the system,  $z_t \in \mathbb{R}^{d_z}$ is the received measurement, $p(x_t|x_{t-1})$ is the transitional density governed by the stochastic process dynamics, $p(z_t|x_{t})$ is the measurement likelihood function, $\mathcal{B}^{-}_t(x_t|z_{1:t-1})$ is the predictive belief distribution at time $t$, and $\mathcal{B}_t(x_t|z_{1:t})$ is the posterior belief of $x_t$ when all measurements $z_{1:t} = [z_1,\ldots,z_t]$ up to time $t$ have been received. Subsequently, given the recursion in Eq. \eqref{eq:Bayes_recursion} the minimum mean square estimator (MMSE) $\hat{x}^{\text{MMSE}}_t$ is given by:
\begin{equation}\label{eq:mmse}
	\hat{x}^{\text{MMSE}}_t = \int x_t \,\mathcal{B}_t(x_t|z_{1:t})\, dx_t.
\end{equation}

However, in our problem, at time \(t\) the observation is a point pattern of random cardinality, not a fixed-length vector, hence the standard vector-likelihood underlying Eq. \eqref{eq:Bayes_recursion} is not directly applicable. We therefore replace the likelihood in Eq.~\eqref{eq:Bayes_recursion} with the appropriate set-likelihood and apply Bayes' rule with that form.
To achieve this, first observe that the transitional density $p(x_t|x_{t-1})$ in Eq. \eqref{eq:Bayes_recursion} becomes:

\begin{equation}
	p(X_t|X_{t-1}) = \int \delta(X_t - \xi(X_{t-1},E_{t-1}))\, P_E(E_{t-1})\, dE_{t-1}, \notag
\end{equation}

\noindent as direct consequence of the fire front stochastic dynamics, where $\delta(\cdot)$ is the Dirac delta function. 
Subsequently, we can compute the predicted belief $\mathcal{B}^{-}_t(X_t|Z_{1:t-1})$ at time step $t$, assuming the posterior density at the previous time step, $\mathcal{B}_{t-1}(X_{t-1}|Z_{1:t-1})$, is known. The posterior belief $\mathcal{B}_t(X_t|Z_{1:t})$ can then be obtained by incorporating the measurement set $Z_t$ via the correction step shown in Eq.~\eqref{eq:Bayes_recursion}, provided that an expression for the measurement likelihood function $p(Z_t|X_t, y_t)$ is available. This process can then be recursively applied to the next time step.

\textbf{Proposition:}
\textit{Let $X_t = [x^1_t,\ldots,x^N_t]^\top$ be the state of the fire front at time $t$, and let
$Z_t=[z^1_t,\ldots,z^{m_t}_t]$ denote the received measurement vector at time step $t$. The likelihood of $Z_t$ given $X_t$ and $y_t$ is given by:} $p(Z_t|X_t,y_t)=$
\begin{equation}\label{eq:joint_likelihood}
\frac{1}{m_t!} \exp\left(-\sum_{i=1}^N \lambda^i_t(x^i_t)\right)
\prod_{k=1}^{m_t} 
\left(\,\sum_{i=1}^N \gamma^i_t\bigl(z_k|x^i_t,y_t\bigr)\right),
\end{equation}

\noindent where $\lambda^i_t(x^i_t) = \int_{O_t} \gamma^i_t(z\,|\,x^i_t, y_t)\, dz$ is the expected number of detections from vertex $i$, $\gamma^i_t(z\,|\,x^i_t, y_t)$ is the Poisson intensity corresponding to vertex $i$ within the sensing region $O_t$ as defined in Eq.~\eqref{eq:PPP}, and $m_t!$ denotes the factorial of $m_t$. Consequently, $p(Z_t | X_t, y_t)$ can be used directly in Eq.~\eqref{eq:Bayes_recursion} enabling the handling of multiple objects and multiple measurements without requiring explicit measurement-to-object association, and allowing the computation of the MMSE as discussed previously.

\begin{proof}
Due to the independence of noise realizations in the measurement process, the measurements are conditionally independent given the fire front state \( X_t \). As a result, the Poisson processes defined by Eq.~\eqref{eq:PPP} are themselves independent, which implies that the combined set of all measurements generated by all processes forms a superposition of Poisson point processes, with total intensity \( \Gamma_t(z|X_t,y_t) = \sum_{i=1}^N \gamma^i_t(z | x^i_t,y_t) \), and probability density function (PDF) \( p(z | X_t) = \Gamma_t(z | X_t,y_t) \left( \int_{O_t} \Gamma_t(z | X_t,y_t) \, dz \right)^{-1} = \Gamma_t(z | X_t,y_t) \Lambda_t^{-1} \).
Subsequently, the joint likelihood \( p(Z_t | X_t,y_t) \) of receiving \( m_t \) measurements at time \( t \) can be decomposed as \( p(Z_t|X_t,y_t) = p_m(m_t) \, p(z^1_t, \ldots, z^{m_t}_t | m_t, X_t, y_t) \), where \( p_m(m_t|y_t) \) is the probability of receiving \( m_t \) observations inside the sensing range, and \( p(z^1_t, \ldots, z^{m_t}_t | m_t, X_t,y_t) \) is the conditional joint likelihood function.
The term \( p_m(m_t,|y_t) \) follows a Poisson distribution with parameter \( \Lambda_t \), i.e., \( p_m(m_t|y_t) = {m_t!}^{-1} \exp(-\Lambda_t) \Lambda_t^{m_t} \). In addition, the joint likelihood decomposes as \( p(z^1_t, \ldots, z^{m_t}_t | m_t, X_t,y_t) = \prod_{k=1}^{m_t} p(z_k | X_t,y_t) \), and thus \( \prod_{k=1}^{m_t} p(z_k | X_t,y_t) = \prod_{k=1}^{m_t} \Gamma_t(z_k | X_t,y_t) \Lambda_t^{-1} \). Since each vertex only contributes locally to its own measurement process, it follows that \( \prod_{k=1}^{m_t} \Gamma_t(z_k | X_t,y_t) \Lambda_t^{-1} = \prod_{k=1}^{m_t} \left( \sum_{i=1}^N \gamma^i_t(z_k | x^i_t,y_t) \Lambda_t^{-1} \right) \). Since $ \int_{O_t} \Gamma_t(z | X_t,y_t)  dz$ = $\sum_{i=1}^N \int_{O_t} \gamma^i_t(z | x^i_t,y_t)  dz$ = $\sum_{i=1}^N \lambda^i_t(x^i_t)$, the result in Eq.~\eqref{eq:joint_likelihood} follows directly.
\end{proof}

\subsection{Information-seeking Predictive Control} \label{ssec:Control}

The problem in Sec.~\ref{ssec:formulation} is addressed via the information-seeking predictive controller shown in Problem~(P1), formulated as a receding horizon SOC problem. The goal is to compute control inputs \( \{u_{t+\tau-1|t}\}_{\tau=1}^{T} \) that optimize the agent's sensing behavior by minimizing the cumulative uncertainty in the (pseudo) posterior beliefs of the fire front states, as defined in Eq.~\eqref{eq:P1_a}. At each time step \( t \), only the first control input \( u_{t|t} \) is applied, and the process is repeated over a shifted horizon.

\begin{figure}[h!]
\begin{subequations}
\begin{align} 
&\hspace*{-3mm}\textbf{Problem (P1):}~\textit{Information-seeking Predictive Control} & \notag\\
&\hspace*{-3mm} \underset{\{u_{t+\tau-1|t}\}_{\tau=1}^{T}}{\min}\mathbb{E}_{Z_t^{1:T}} \left\{ \sum_{\tau=1}^{T} \nu^{\tau} \mathcal{C}_{\tau}\! \left( \mathcal{B}_{t+\tau|t}(\cdot|Z_{1:t+\tau|t}) \right) \right\} &   \label{eq:P1_a} \\
&\hspace*{-3mm}\text{subject to:  } ~  &\nonumber\\
&\hspace*{-3mm}\mathcal{B}^{-}_{t+\tau|t} = \int p(X_{\tau}|X_{\tau-1}) \, \mathcal{B}_{t+\tau-1|t}(X_{\tau-1}|\cdot) \, dX_{\tau-1}, & \hspace*{-25mm}  \label{eq:P1_b}\\
&\hspace*{-3mm}\mathcal{B}_{t|t}= \mathcal{B}_{t|t-1}, & \hspace*{-25mm} \label{eq:P1_c}\\
&\hspace*{-3mm} y_{t+\tau|t} = f_a(y_{t+\tau-1|t}, u_{t+\tau-1|t}), & \hspace*{-25mm}  \label{eq:P1_d}\\
&\hspace*{-3mm} y_{t|t} = y_{t|t-1}, & \hspace*{-25mm} \label{eq:P1_e}\\
&\hspace*{-3mm} \mathcal{B}_{t+\tau|t}(\cdot|Z_{1:t+\tau|t}) \propto p\left(Z_{t+\tau|t}|X_{\tau},y_{t+\tau|t}\right)\mathcal{B}^{-}_{t+\tau|t},  & \hspace*{-25mm}  \label{eq:P1_f}\\
&\hspace*{-3mm} y_t \in \mathcal{Y}, u_t \in \mathcal{U}, X_t \in \mathcal{X}, Z_t \in \mathcal{Z}, & \label{eq:P1_g} \\
&\hspace*{-3mm} \mathcal{C}_{\tau} \in [0,1], \nu \in (0,1], \tau=\{1,\ldots,T\}.  & \label{eq:P1_h}
\end{align}
\end{subequations}
\vspace{-5mm}
\end{figure}

For each prediction time step $\tau \in \{1,\ldots,T\}$ in the horizon, the agent predicts the fire front state $X_{\tau}$ forward in time using the Bayesian prediction step, based on the transition density $p(X_{\tau} | X_{\tau-1})$ and the (pseudo) posterior belief from the previous time step, $\mathcal{B}_{t+\tau-1|t}(X_{\tau-1} | Z_{1:t+\tau-1|t})$, as shown in Eqs.~\eqref{eq:P1_b}-\eqref{eq:P1_c}. The constraints in Eqs.~\eqref{eq:P1_d}-\eqref{eq:P1_e} arise from the agent dynamics, which are assumed to be deterministic in this work. At the predicted time step $\tau$, given the agent's predicted state $y_{t+\tau|t}$ and sensing range $O_{t+\tau|t}$, the agent receives the predicted measurement set $Z_{t+\tau|t}$. This set is then used to compute the posterior belief $\mathcal{B}_{t+\tau|t}(X_{\tau} | Z_{1:t+\tau|t})$ via the Bayesian correction step, as shown in Eq.~\eqref{eq:P1_f}, using the joint likelihood function $p(Z_{t+\tau|t} | X_{\tau}, y_{t+\tau|t})$ and the predicted density. This posterior distribution subsequently becomes the prior for the next prediction step, continuing the recursive process.
The predicted measurements \( Z_{t+\tau|t} \) represent hypothetical observations based on the predicted control inputs and the anticipated fire front state. Since actual measurements are only available after executing the control actions, the objective in Eq.~\eqref{eq:P1_a} requires taking an expectation over all possible future measurement sequences. This enables informed decision-making by accounting for potential outcomes without executing the corresponding trajectories.

Equations \eqref{eq:P1_b}-\eqref{eq:P1_f} admit no closed form: the model is nonlinear and non-Gaussian with set-valued (multi-object, multi-measurement) observations, so Kalman-type filters are inapplicable. 
Therefore, the recursion is implemented using Sequential Importance Resampling (SIR), i.e., particle filtering. Specifically, the belief $\mathcal{B}_\tau$ is represented by a set of weighted particles $\mathcal{B}_\tau=\{w^{(i)}_\tau, X^{(i)}_\tau\}_{i=1}^{N_s}$, where $X^{(i)}_\tau = [x^1_\tau, \ldots, x^N_\tau]^\top$. These particles are propagated to the next time step according to the process dynamics and reweighted using the likelihood function to compute the posterior. For notational convenience, we will often write $t+\tau|t$ as $\tau$ when no ambiguity arises. The functional $\mathcal{C}_{t+\tau|t} : \mathcal{B}_{t+\tau|t}(X_{t+\tau|t}|Z_{1:t+\tau|t}) \rightarrow [0,1]$ in Eq.~\eqref{eq:P1_a} quantifies the uncertainty of the fire-front state $X_{t+\tau|t}$ encoded in the posterior distribution at time $t+\tau|t$, conditioned on all hypothetical measurements $Z_{1:t+\tau|t}$. This uncertainty is measured by the \emph{Risk-Weighted Dispersion} (RWD), defined as:
\begin{equation}\label{eq:rwd}
\mathcal{C}_{t+\tau|t}\!\big(\mathcal{B}_{t+\tau|t}\big)
= \frac{1}{\omega}\sum_{\varepsilon \in \tilde{\mathcal{E}}} \mathcal{R}(\varepsilon)\, \det\!\big(\Sigma^\varepsilon_{t+\tau|t}\big),
\end{equation}
\noindent where the environment $\mathcal{E}$ is discretized in space to form a 2D grid $\tilde{\mathcal{E}}$, composed of a finite number of non-overlapping, equally sized cells $\tilde{\mathcal{E}} = \{\varepsilon_1, \ldots, \varepsilon_{|\tilde{\mathcal{E}}|}\}$, such that $\bigcup_{i=1}^{|\tilde{\mathcal{E}}|} \varepsilon_i = \mathcal{E}$. The term $\mathcal{R}(\varepsilon) \in [0,1]$ denotes the risk value associated with cell $\varepsilon$, reflecting the severity of fire presence in that region. The random quantity $\det(\Sigma_{t+\tau|t}^\varepsilon)$ is the determinant of the sample covariance matrix $\Sigma_{t+\tau|t}^\varepsilon$ computed from all particle points residing in cell $\varepsilon$ at time step $t+\tau|t$, and $\omega$ is a scaling factor ensuring that $\mathcal{C}_{t+\tau|t}\!\big(\mathcal{B}_{t+\tau|t}\big)\in[0,1]$.

 In practice, the computation of the aggregate cell dispersion $\det\!\big(\Sigma^\varepsilon_{t+\tau|t}\big)$ is implemented via a vertex-centric pooling operation. For each fire-front vertex $i \in \{1,\dots,N\}$, we explicitly compute its $2\times 2$ covariance matrix $\Sigma^i_{t+\tau|t}$ across its entire ensemble of $N_s$ particle realizations, which captures the full spatial dispersion of the vertex. Subsequently, the cell dispersion term $\det\!\big(\Sigma^\varepsilon_{t+\tau|t}\big)$ in Eq.~\eqref{eq:rwd} is evaluated operationally by summing the individual uncertainty volumes, i.e., $\sum_i \det\!\big(\Sigma^i_{t+\tau|t}\big), $ over all vertices whose expected spatial location (MMSE) resides within cell $\varepsilon$.

Finally, the parameter $\nu \in (0,1]$ in Eq. \eqref{eq:P1_a} is a discount factor that controls the relative importance of future decisions.

\subsection{Adaptive LCB-guided Policy Search} \label{ssec:Search}

Problem (P1) is a stochastic, multi-dimensional, non-linear, and non-convex optimization problem that cannot be directly solved in its original form. However, we can address an equivalent version by reformulating (P1) as a Markov Decision Process (MDP) \cite{Sutton98}. To achieve this, we assume that the agent's control inputs $u_t \in \mathcal{U}$ can be reduced to a finite set $\mathbb{U}$, consisting of $|\mathbb{U}|$ discrete control vectors $\hat{u}_t \in \mathbb{U}$. This discretization, in turn, leads to a finite set of possible agent states $\hat{y}_t \in \mathbb{Y}_t$. Consequently, (P1) can be reformulated as an MDP $\langle \mathcal{S}, \mathbb{U}, \mathcal{T}, \mathcal{C} \rangle$, where $\mathcal{S}$ is the state space of the system, and an individual state $s \in \mathcal{S}$ is represented as the tuple $s_t = (\mathcal{B}_t, \hat{y}_t)$. Note that the fire front process evolves independently of the agent's actions. The transition function $\mathcal{T} : \mathcal{S} \times \mathbb{U} \rightarrow \mathcal{S}$ describes the evolution of the system in response to agent actions. Although the agent's actions are deterministic in our setting, the transition function $\mathcal{T}$ remains stochastic due to the randomness in the agent's observations upon executing an action. Specifically, we have: $\mathcal{T} : p\left(s^\prime_t = (\mathcal{B}^\prime, \hat{y}^\prime) \mid s_{t-1} = (\mathcal{B}, \hat{y}), \hat{u}_{t-1} = u\right)$. The cost function $\mathcal{C}$ assigns a cost to a specific state $s^\prime$ resulting from applying action $\hat{u}$ at state $s$. With slight abuse of notation, we denote it as $\mathcal{C}(s^\prime = (\mathcal{B}, \hat{y}))$, which effectively operates on the posterior belief $\mathcal{B}$ in state $s^\prime$, derived from the agent state $\hat{y}$. The cost is defined according to Eq. \eqref{eq:rwd}.

\begin{figure*}
	\centering
	\includegraphics[width=\textwidth]{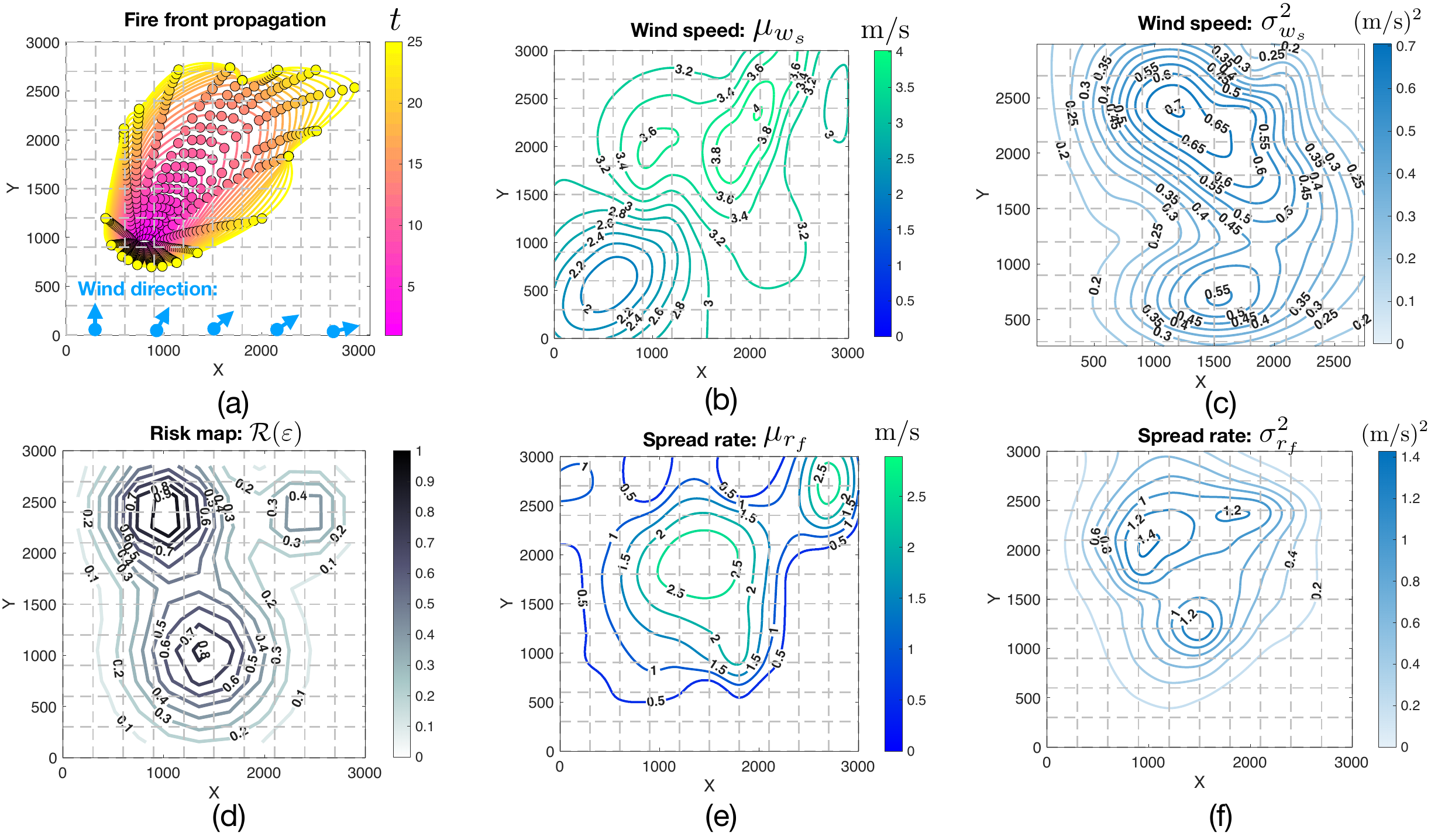}
	\caption{Simulation Setup: (a) Fire front true evolution, (b)(c) Wind speed parameters, (d) Risk map, and (e)(f) Fire spread rate parameters.}
	\label{fig:fig1}
	\vspace{-0mm}
\end{figure*}

We define a finite-horizon open-loop policy over $T$ steps as the control input sequence $\pi= \{\hat{u}_{t+\tau-1|t}\}_{\tau=1}^{T} \in \mathbb{U}^T$. Let $\Pi_T=\mathbb{U}^T$ denote the set of all such admissible policies. Each policy is a predetermined sequence of actions evaluated via simulation of the MDP's stochastic state transitions. Given that the agent starts from an initial state $s_t = (\mathcal{B}_t, \hat{y}_t)$ at time step $t$, a policy $\pi \in \Pi_T$ can then be simulated under the MDP to obtain:
\begin{equation}\label{eq:value_function}
	V^{\pi}_t(s_t) = \mathbb{E} \left\{ \sum_{\tau=1}^{T} \nu^{\tau} \mathcal{C}_{\tau}\left( s^{\pi}_{t+\tau|t} \right) \right\} ,
\end{equation}
\noindent where $s^{\pi}_{t+\tau|t}$ is the state encountered at level $\tau$ in the horizon under policy $\pi$ when starting from state $s_t$. Subsequently, the optimal policy that minimizes the expected cumulative cost over the horizon is given by 
\begin{equation}\label{eq:MDP_best_policy}
	\pi^\star = \arg\min_{\pi \in \Pi_T} \, V^{\pi}_t(s_t)
\end{equation}
\noindent Observe that the optimization problem in Eq.~\eqref{eq:MDP_best_policy} is equivalent to Problem (P1) under the assumption of a finite number of admissible control inputs. Solving this equivalent MDP formulation however, introduces its own set of challenges. In particular, the continuous belief-space represented by particles and the stochastic nature of the measurements result in an effectively infinite state space. This makes classical dynamic programming (DP) methods, such as value iteration and backward induction, infeasible. These approaches rely on a finite state space to compute value functions backward from the planning horizon and therefore do not apply here. Furthermore, the transition dynamics are not explicitly known in closed form, nor can they be represented in a tabular format. 

To solve this problem we build upon the Upper Confidence Bound 1 (UCB1) framework \cite{Auer2002finite} and we treat each sequence $\pi = \{\hat{u}_{t+\tau-1|t}\}_{\tau=1}^{T} \in \mathbb{U}^T$ as an \textit{arm} in a multi-armed bandit. We use rollouts to simulate policy outcomes and employ a UCB1-like adaptive selection strategy to efficiently balance exploration and exploitation. In particular, the UCB1 algorithm iteratively computes an upper confidence bound score on the expected reward for each arm (where arms are considered the actions), by adding the sample mean reward of the arm to an exploration bonus that depends on both the total number of arm pulls and the number of pulls for that specific arm. 
\begin{algorithm}
\caption{\textbf{Adaptive LCB-guided Search}}
\begin{algorithmic}[1]
\State \textbf{Input:} Policy set $\Pi_T$, Initial state $s_t=(\mathcal{B}_t,\hat{y}_t)$\State Initialize $I^0(\pi)=0, \forall \pi \in \Pi_T$
\For{$n = 1, \dots, (n_{\max}\geq|\Pi_T|)$}
\State Sample policy: $\!\hat{\pi} \!=\! \arg\min \mathrm{LCB}^{n-1}(\pi,s_t)$, Eq. \eqref{eq:LCB}
\State Initialize rollout cost: $c_0 = 0$
	\For{$\tau = 1, \dots, T$} 
	\State Compute predictive density: $\mathcal{B}^{-}_{t+\tau|t}$ via Eq. \eqref{eq:P1_b}
	\State Move to new state: $\hat{y}_{t+\tau|t} = f_a\left(\hat{y}_{t+\tau-1|t},\hat{\pi}_\tau\right)$
	\State Sample latent state: $\tilde{X}_{t+\tau|t} \sim \mathcal{B}^{-}_{t+\tau|t}$
	\State Sample meas.: $Z_{t+\tau|t} \sim p(Z|\tilde{X}_{t+\tau|t},\hat{y}_{t+\tau|t})$
	\State Compute posterior: $\mathcal{B}_{t+\tau|t}$ via Eq. \eqref{eq:P1_f}
	\State Compute stage cost: $c_\tau = c_{\tau-1} +  \nu^{\tau} \mathcal{C}_\tau(\mathcal{B}_{t+\tau|t})$
	\EndFor
	\State Update: $Q^n(\hat{\pi},s_t) \!=\! \mathrm{mean}\left(Q^{n-1}(\hat{\pi},s_t), \frac{c_T}{L_{\max}}\right)$
	\State \>\>\>\>\>\>\>\>\>\>\>\>\>\>\>\ $I^n(\hat{\pi}) = I^{n-1}(\hat{\pi}) + 1$
\EndFor
\State \textbf{Output:} Optimal policy $\pi^\star = \arg \min~ Q^{n_\text{max}}(\pi,s_t)$
\end{algorithmic}
\end{algorithm}
This formulation is grounded in the ``optimism under uncertainty'' principle and leverages concentration inequalities (i.e., Chernoff-Hoeffding bounds) to guarantee near-optimal regret (i.e., the expected loss between the optimal policy and selected policies). Based on these scores, the policy always selects the arm with the highest current upper confidence bound, which enables asymptotic convergence to the optimal policy \cite{Auer2002finite}.

Specifically, the proposed adaptive-search algorithm iteratively computes a Lower Confidence Bound (LCB) on the expected total cost associated with each $T$-finite control sequence, which is then used to adaptively select the next policy to simulate. Specifically, for an agent at state $s_t$, we define the LCB score of a policy $\pi \in \Pi_T$ at iteration $n$ as:

\begin{equation}\label{eq:LCB}
	\mathrm{LCB}^{n-1}(\pi,s_t)\!\! =\!\!
	\begin{cases}
	\!\! Q^{n-1}(\pi,s_t)\! - \!\sqrt{\frac{2 \ln n}{I^{n-1}(\pi)}},\!\!\! &  I^{n-1}(\pi) > 0, \\
	\!\!\mathrm{LCB}^{n-1}_{\min},\!\!\! & \text{otherwise},
	\end{cases}
\end{equation}

\noindent where $I^{n-1}(\pi)$ denotes the number of times policy $\pi$ has been simulated prior to iteration $n$, and $Q^{n-1}(\pi,s_t) = \frac{1}{I^{n-1}(\pi)} \sum_{i=1}^{n-1} \mathds{1}_{\{\hat{\pi}^i = \pi\}} L^{(i)}(\pi,s_t)$ is the empirical sample mean of the cumulative cost incurred by policy $\pi$ up to iteration $n-1$. Here, $\hat{\pi}^i$ is the policy selected at iteration $i$, $\mathds{1}_{\{\cdot\}}$ is the indicator function, and $L^{(i)}(\pi,s_t) = \frac{1}{L_{\max}} \sum_{\tau=1}^{T} \nu^{\tau} \mathcal{C}_{\tau}\!\left(s^{\pi, (i)}_{t+\tau|t}\right)$ is the normalized total discounted cost realized explicitly during the $i$-th simulated rollout of policy $\pi$ starting from state $s_t$, with $L_{\max} = \sum_{\tau=1}^{T} \nu^{\tau}$. The constant $\mathrm{LCB}^{n-1}_{\min} < -\sqrt{2 \ln n}$ ensures that any policy not yet selected by iteration $n$ will have an LCB value lower than that of any previously selected policy, thereby guaranteeing it will be explored with higher probability in subsequent iterations. Finally, the term $\sqrt{\frac{2 \ln n}{I^{n-1}(\pi)}}$ serves as an exploration bonus, encouraging the selection of under-explored policies. 


The complete LCB-guided adaptive search algorithm is shown in Algorithm~1. At each time step $t$, the algorithm identifies the optimal policy $\pi^\star$ over the horizon $\{t+\tau|t\}_{\tau=1}^{T}$. The agent then executes the first control input of $\pi^\star$, transitions to a new state, receives the ``real'' measurement, computes the posterior belief $\mathcal{B}_{t+1}$, and the algorithm is re-applied at the next time step $t + 1$. In Note that in Alg.~1, $\mathrm{mean}\!\big(Q^{n-1}(\hat{\pi},s_t), \frac{c_T}{L_{\max}}\big)$ stands for the recursive empirical average update, computed as $Q^{n-1}(\hat{\pi},s_t) + \frac{1}{I^n(\hat{\pi})}\!\left(\frac{c_T}{L_{\max}} - Q^{n-1}(\hat{\pi},s_t)\right)$. Finally, in Alg.~1, the predicted measurements are simulated via ancestral sampling i.e., a latent fire-front state is first drawn from the predictive belief $\tilde{X}_{t+\tau|t} \sim \mathcal{B}^{-}_{t+\tau|t}$, and the measurement set is subsequently generated from the measurement likelihood conditioned on this latent state and the planned agent state. We should note that intelligent pruning techniques, such as $\epsilon$-suboptimal reductions~\cite{Atanasov2014information}, can be designed and integrated into the proposed approach to focus on the most promising set of control inputs at each time step thereby reducing the runtime complexity.

\textbf{Theorem 1 (Completeness):} \textit{Let $\Pi_T$ be the set of all candidate policies in the LCB framework. Then for every optimal policy $\pi^\star \in \Pi_T$, there exists iteration $n$ such that $\pi^\star$ is selected at least once by the algorithm.}

\begin{proof}
By construction of the LCB algorithm, each policy $\pi \in \Pi_T$ is initialized in a manner that guarantees it is tried at least once. Thus, any policy that has never been tried up to a certain point will have a lower LCB score than those policies with a finite sample count, ensuring it is selected at least once. Hence $\pi^\star$ is included among the solutions eventually tried.
\end{proof}

\textbf{Theorem 2 (Asymptotic Convergence):}
\textit{Suppose there is a unique optimal policy $\pi^\star$ whose expected cost is strictly smaller than that of any other $\pi \in \Pi_T$. Under LCB, the expected average regret decays on the order of $O \bigl(\tfrac{\ln n}{n}\bigr)$, and the fraction of times $\pi^\star$ is selected converges to $1$ as $n \to \infty$.}

\begin{proof} 
By Theorem~1, every policy in $\Pi_T$ is selected at least once under LCB. Moreover, LCB follows the same selection rule as UCB1 for minimization problems, adapted to entire control sequences, effectively treating each policy as a bandit arm with a stationary cost distribution and stage costs bounded in the range $[0, 1]$. Consequently, the classical finite-time analysis on UCB1 \cite{Auer2002finite} applies directly: the expected number of times any suboptimal policy is selected is bounded by $O(\ln n)$. Hence, the expected fraction of times a suboptimal policy is chosen decreases as $O \bigl(\frac{\ln n}{n}\bigr)$, which implies the expected average regret shrinks at that same rate. Thus, as $n \to \infty$, the average regret approaches $0$ and the fraction of times the optimal policy $\pi^\star$ is selected asymptotically converges to 1. 
\end{proof}

%

\section{Evaluation} \label{sec:Evaluation}
\begin{figure*}
	\centering
	\includegraphics[width=\textwidth]{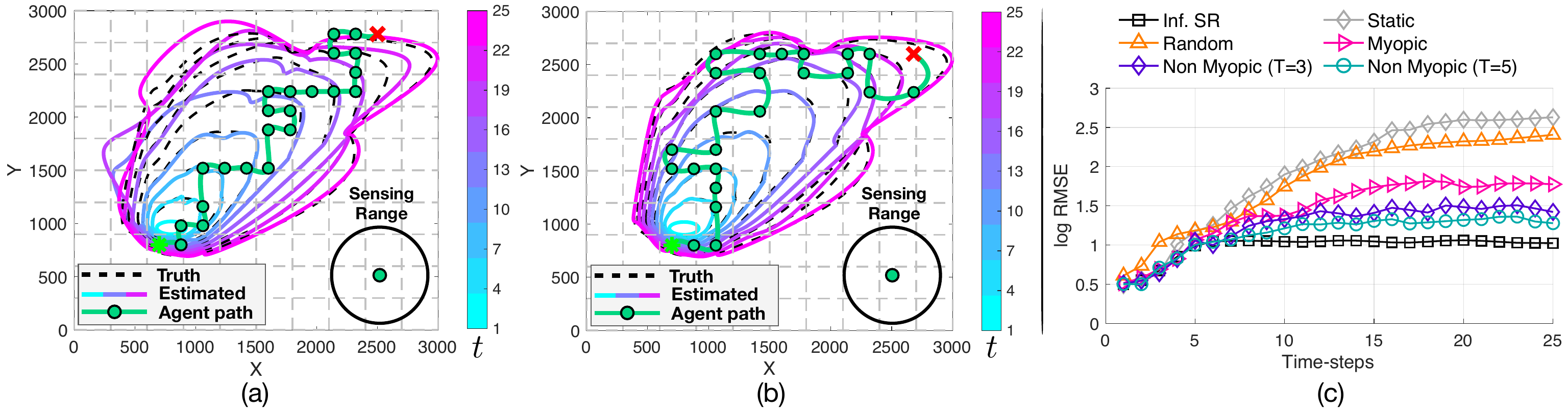}
	\caption{Performance Evaluation: (a) Agent trajectory ($\star$ and $\times$ indicate start and stop states respectively), and fire front state (i.e., estimated with solid line, and true state with dotted line) obtained with Alg. 1, over 25-steps experiment, with myopic settings (i.e., horizon $T=1$), (b) Same result with non-myopic settings (i.e. horizon $T=3$), (c) Estimation error for different configurations of the proposed approach and comparisons with baselines.}
	\label{fig:fig2}
	\vspace{-0mm}
\end{figure*}

 \subsubsection{Simulation Setup}
 To evaluate the proposed approach, we used the following setup: the environment $\mathcal{E} \subset \mathbb{R}^2$ is square, with side length $3 \times 10^3~\mathrm{m}$ in each dimension, whereas its discrete representation $\tilde{\mathcal{E}}$ consists of a $10 \times 10$ grid with equally sized cells, as shown in Fig.~\ref{fig:fig1}(a) with gray dotted lines. The fire front state $X_t$ comprises $N = 20$ fire front vertices (shown in Fig.~\ref{fig:fig1}(a)), initially forming an ellipse centered at $(x, y) = (800, 900)$, i.e., the ignition point, with semi-major and semi-minor axes of lengths $120~\mathrm{m}$ and $60~\mathrm{m}$, respectively. Each of these vertices evolves according to Eq.~\eqref{eq:fire_model_vertex} with $\Delta t = 60\mathrm{s}$. The environmental conditions are as follows: wind direction $\theta$, wind speed $w_s$, and fire spread rate $r_f$ due to fuel are defined for each cell $\varepsilon \in \tilde{\mathcal{E}}$. Specifically, the mean wind direction (with North aligned with the $y$-axis) varies uniformly across the $x$-axis from North to North-East to East, as illustrated by the blue arrows in Fig.~\ref{fig:fig1}(a), following a von Mises distribution with concentration parameter $\kappa = 500$ in every cell i.e., $\theta(\varepsilon) \sim \mathcal{VM}(\mu_{\theta(\varepsilon)}, \kappa_{\theta(\varepsilon)})$. Subsequently, the wind speed and fire spread rate are modeled as rectified Gaussian distributions, i.e., $w_s(\varepsilon) \sim \mathcal{N}_R(\mu_{w_s(\varepsilon)}, \sigma^2_{w_s(\varepsilon)})$ and $r_f(\varepsilon) \sim \mathcal{N}_R(\mu_{r_f(\varepsilon)}, \sigma^2_{r_f(\varepsilon)})$. The superposition of these random variables over the grid for each $\varepsilon \in \tilde{\mathcal{E}}$ is shown in Fig.~\ref{fig:fig1}(b) and Fig.~\ref{fig:fig1}(c) for the wind speed, and in Fig.~\ref{fig:fig1}(e) and Fig.~\ref{fig:fig1}(f) for the fire spread rate, respectively. 
 Finally, the risk $\mathcal{R}(\varepsilon)$ associated with each cell, indicating the severity of fire in that region is shown in Fig.~\ref{fig:fig1}(d).  Consequently, the fire front state $X_t$ evolves in continuous space and the environmental parameters $\theta$, $w_s$, and $r_f$ influencing each fire front vertex are obtained by associating it with the nearest cell in the discretized environment. The mobile agent (i.e., a drone) evolves according to 
$\hat{y}_t = \hat{y}_{t-1} + d_R \Delta t \bigl[\cos_d(\vartheta), \sin_d(\vartheta)\bigr]^\top$, and is controlled via the input $\hat{u}_t = [d_R \Delta t,\ \vartheta]$, where $d_R \in \{3,6\}~\mathrm{m/s}$ and $\vartheta \in \{0, 90, 180, 270\}~\mathrm{deg}$. We assume that the drone operates at a fixed altitude of $250~\mathrm{m}$ and is equipped with a wide-angle field-of-view camera with a viewing angle of $120~\mathrm{deg}$, resulting in a circular sensing range with radius $R_a = 250 \tan_d(120/2) \approx 430~\mathrm{m}$. The measurement noise is set to $\sigma_z = 3.5~\mathrm{m}$, and we assume a fixed intensity $\lambda^i_t(x^i_t) = 5$, for all $i \in \{1, \ldots, N\}$ and all $t$. Finally, $\nu=0.99$, $n_\text{max}$ varies depending on the horizon, and the Bayes recursion in Eq.~\eqref{eq:Bayes_recursion} is implemented as a SIR particle filter with $N_s = 2000$ particles.

\subsubsection{Results}

Figure~\ref{fig:fig1}(a) illustrates the true evolution of the fire front over a simulation period of $T_s = 25$ time steps, influenced by the environmental conditions described above. The fire-spread physics in this model support propagation in all directions from the ignition point. As a result, even under strong wind and fuel conditions, the upwind (back) perimeter of the fire continues to advance, albeit at a slower rate, as depicted. Furthermore, Fig.~\ref{fig:fig1}(a)-(c) illustrates how variations in wind direction, wind speed, and spread rate influence the propagation of the fire front, either by accelerating or decelerating its advancement.

The output of Alg.~1 for this setup is shown in Fig.~\ref{fig:fig2}(a) and Fig.~\ref{fig:fig2}(b), corresponding to horizon lengths of $T=1$ (myopic) and $T=3$ (non-myopic), respectively. The agent is initialized at position $(x, y) = (700, 800)$, as indicated by the green asterisk, running Alg.~1 in a rolling horizon fashion, as discussed in Sec.\ref{ssec:Search}. The green line in the figures is the agent's final trajectory over 25 time steps, resulting from the minimization of the expected cumulative RWD over the planning horizon at each time step. The dotted black lines show the true evolution of the fire front, while the time color-coded lines represent the estimated front; ideally, these two should align closely.
As shown, the non-myopic approach plans multiple steps ahead and achieves improved performance compared to the myopic strategy, which plans greedily. Specifically, the myopic behavior of the agent in Fig.~\ref{fig:fig2}(a) prevents it from targeting the high-risk region in the top-left corner of Fig.~\ref{fig:fig1}(d), which in this scenario is also associated with high uncertainty, as illustrated in Figs.~\ref{fig:fig1}(c) and \ref{fig:fig1}(f). This limitation results in significant estimation errors, as shown.

Subsequently, Fig.~\ref{fig:fig2}(c) illustrates the performance of the proposed approach in terms of root mean square error (RMSE) i.e., $\sqrt{\mathbb{E}\big((\hat{X}_t - X_t)^2\big)}$, between the estimated fire front state $\hat{X}_t$ and the true state $X_t$ over 25 time steps, using a uniform risk map. Specifically, we conducted 50 Monte Carlo trials, randomly initializing both the fire front and the agent's position within the simulation environment described earlier.
The figure presents the average $\log_{10} \mathrm{RMSE}$ per time step per vertex over the 25-step simulation, comparing six different approaches. The baseline, denoted as \textit{Inf. SR}, corresponds to an agent with infinite sensing range that remains stationary and simply runs the particle filter. In this case, the RMSE arises solely from measurement noise and multiple detections, with no influence from control actions-representing the best achievable performance under the given settings.
The \textit{Static} approach involves a stationary agent with a finite sensing range, while the \textit{Random} approach uses an agent also with finite sensing range that selects a random control input at each time step. As expected, both approaches result in significant errors. The figure also includes the proposed method evaluated with different planning horizon lengths, i.e., $T = 1$ (myopic), and non-myopic settings with $T = 3$ and $T = 5$, and clearly demonstrates improved performance as the planning horizon increases. Note that the baseline is unattainable in this setting due to the agent's limited sensing capabilities.

\vspace{-0mm}
\section{Conclusion} \label{sec:conclusion}

This paper considers the problem of fire front monitoring under uncertainty by formulating it as a stochastic optimal control problem that integrates sensing, estimation, and control. A recursive Bayesian estimator was developed for elliptical-growth fire front processes, and the control problem was formulated as a finite-horizon Markov Decision Process (MDP). An information-seeking control law was then derived using a lower confidence bound (LCB)-based adaptive search, enabling optimal risk-aware planning.

\flushbottom
\balance

\bibliographystyle{IEEEtran}
\bibliography{IEEEabrv,main}

\end{document}